\documentclass[a4paper,fleqn,longmktitle]{cas-sc}

\usepackage[authoryear]{natbib}
\usepackage{url}
\usepackage{booktabs}
\usepackage{xcolor, colortbl}
\usepackage{graphicx}
\usepackage{tabularx}
\usepackage{etoolbox}
\usepackage{siunitx}
\usepackage{tipa}
\usepackage[noabbrev, capitalize, nameinlink]{cleveref}
\usepackage{multirow}

\definecolor{jmcolor}{RGB}{255,127,0}

\newcommand{\ignore}[1]{}

\newcommand{\seqten}{\cellcolor[HTML]{0d0982}{\textcolor[HTML]{f1f1f1}{$10$}}}
\newcommand{\seqthirty}{\cellcolor[HTML]{2e0595}{\textcolor[HTML]{f1f1f1}{$30$}}}
\newcommand{\seqhundred}{\cellcolor[HTML]{7e03a8}{\textcolor[HTML]{f1f1f1}{$100$}}}
\newcommand{\seqtwoseventy}{\cellcolor[HTML]{f68f44}{\textcolor[HTML]{000000}{$270$}}}
\newcommand{\seqyear}{\cellcolor[HTML]{f0f921}{\textcolor[HTML]{000000}{$365$}}}
\usepackage{fancyhdr}
\fancypagestyle{firststyle}
{
   \fancyhf{}
   \fancyfoot[L]{\footnotesize Gauch et al. (Preprint)}
   \fancyfoot[R]{\footnotesize Page \thepage}
}
\fancypagestyle{otherstyle}
{
   \fancyhf{}
   \fancyhead[C]{\footnotesize The Proper Care and Feeding of CAMELS}
   \fancyfoot[L]{\footnotesize Gauch et al. (Preprint)}
   \fancyfoot[R]{\footnotesize Page \thepage}
}
\begin{document}
\let\WriteBookmarks\relax
\def\floatpagepagefraction{1}
\def\textpagefraction{.001}
\shorttitle{The Proper Care and Feeding of CAMELS}
\shortauthors{Gauch et~al.}
\title[mode = title]{The Proper Care and Feeding of CAMELS: How Limited Training Data Affects Streamflow Prediction}

\author[1]{Martin Gauch}[auid=000, orcid=0000-0002-4587-898X]
\ead{martin.gauch@uwaterloo.ca}
\cormark[1]
\author[2]{Juliane Mai}[auid=001, orcid=0000-0002-1132-2342]
\author[1]{Jimmy Lin}[auid=002]
\cortext[cor1]{Corresponding author}
\address[1]{David R.\ Cheriton School of Computer Science, University of Waterloo, ON, Canada}
\address[2]{Civil and Environmental Engineering, University of Waterloo, ON, Canada}


\begin{abstract}
Accurate streamflow prediction largely relies on historical meteorological records and streamflow measurements.
For many regions, however, such data are only scarcely available.
Facing this problem, many studies simply trained their machine learning models on the region's available data, leaving possible repercussions of this strategy unclear.
In this study, we evaluate the sensitivity of tree- and LSTM-based models to limited training data, both in terms of geographic diversity and different time spans.
We feed the models meteorological observations disseminated with the CAMELS dataset, and individually restrict the training period length, number of training basins, and input sequence length.
We quantify how additional training data improve predictions and how many previous days of forcings we should feed the models to obtain best predictions for each training set size.
Further, our findings show that tree- and LSTM-based models provide similarly accurate predictions on small datasets, while LSTMs are superior given more training data.
\end{abstract}


\maketitle
\thispagestyle{firststyle}

\section{Introduction}

Accurate streamflow predictions are an indispensable prerequisite for water management and flood forecasting.
There is a long history of research that employs machine-learning techniques to model streamflow, and a number of studies have shown that data-driven techniques can outperform traditional approaches based on physical models~\citep{Dawson1998ANN, Dibike2001ANN, Best2015Benchmark, Gauch2019DataDriven, Kratzert2019Benchmark}.
Accurate predictions given scarce training data are especially important, as access to historical records of streamflow and meteorological measurements are exceedingly limited in many regions of the world.
Even in the United States, the number of streamflow gauging stations is on the decline \citep{Fekete2015InSitu}.

Data-driven models---particularly machine learning models---are well-known to be data-hungry and produce more accurate models the more data they are fed~\citep{Amari1993Learning, Banko2001Size}.
In this study, we focus on two types of machine learning models: tree-based models and Long Short-Term Memory (LSTM)-based neural networks.
LSTMs are a machine learning architecture for time-series prediction~\citep{Hochreiter1997LSTM}.
Intuitively, they take a time series, e.g., of forcings, as input and update internal memory states at each time step.
Based on the input and the internal states, the LSTM calculates an output value.
The way these internal state and output values are calculated is governed by a set of parameters, or weights, that are learned during training (hydrologists would refer to this training process as calibration).
For a more detailed description of LSTMs in the context of streamflow prediction, we refer to \citet{Kratzert2018LSTM}, who also introduce the Entity-Aware LSTM (EA-LSTM) architecture we use in this study.
EA-LSTMs are an adaption of LSTMs that take static basin attributes and time-series forcings as separate inputs, so we do not need to concatenate the static attributes to each forcing time step.

Tree-based models, on the other hand, are not explicitly designed for time-series prediction, as they take a flat vector of variables as input; hence, multiple input time series need to be concatenated into a single vector.
The training (calibration) procedure constructs a tree structure, where every leaf node corresponds to a predicted value.
Every path to a leaf corresponds to a conjunction of input features that results in the prediction of the leaf value.
To avoid overfitting---the model learning the training data by memorization rather than learning actual input--output-relations---regression trees usually employ regularization methods that limit the model's expressiveness, e.g., by imposing a maximum tree depth.
More sophisticated tree-based methods, such as gradient-boosted regression trees (GBRT), train multiple trees sequentially to correct the remaining error of the previous trees~\citep{Friedman2001Boosting}.
The overall prediction is then calculated as the sum of each tree's individual prediction.
An example for a GBRT-based model is XGBoost~\citep{Chen2016XGB}, which we employ in this study.

Not all data-driven techniques exhibit the same degree of data demand:\ while neural networks generally need a lot of training data to yield accurate predictions~\citep{Hestness2017Scaling}, tree-based models have been shown to work well when provided with limited training data in certain domains~\citep{Gauch2019Limited}. 
In light of this tension between a data-scarce problem and data-hungry algorithms, our study aims to provide guidance for two particular types of data-driven models, tree- and LSTM-based architectures.

While LSTMs as time-series prediction models are clearly a suitable choice for the problem at hand, tree-based models might seem like a poor fit for this task due to the requirement of flat input vectors.
In recent years, however, researchers and practitioners have used gradient-boosted regression trees (GBRT) with great success in numerous time-series prediction tasks~\citep{Luo2019KDD, Kaggle2016Grupo, Kaggle2015Rossmann1, Kaggle2016Rossmann2}.
Related to streamflow prediction, \citet{Gauch2019Limited} found the GBRT-framework XGBoost to provide more accurate predictions than LSTMs.
In a similar study, \citet{Kratzert2019Benchmark} successfully applied LSTM-based architectures to streamflow prediction.
The two studies, however, used different datasets:
While \citet{Gauch2019Limited} used gridded forcing data for 46~watersheds in the Lake Erie region, \citet{Kratzert2019Benchmark} employed 531~watersheds from the far larger CAMELS dataset~\citep{Newman2014Camels, Addor2017CAMELS}, which range across the continental United States.
The CAMELS dataset also includes basin-averaged forcings for each watershed. 
Hence, the results of both studies cannot be directly compared, and it remains unclear how the predictive quality of tree- and LSTM-based models compares on data spanning different regions and time periods.

In this study, we use parts of the CAMELS dataset and compare tree- and LSTM-based models, answering the following two questions:

\begin{itemize}
 \item When we train a model on a given number of training years, we can still choose how many previous days of forcings we feed the model to generate a single prediction---the input sequence length. Which sequence length yields the best predictions for each model, and does this depend on the training set size?
 \item How do training period length and the number of basins in the training dataset affect the prediction quality of tree- and LSTM-based models?
\end{itemize}

To answer these questions, we first briefly describe the datasets and methods we use (\cref{sec:Data_and_Methods}). 
We present our results in \cref{sec:Results} and conclude with a discussion and an outlook towards future research in \cref{sec:Discussion}.

\section{Data and Methods}
\label{sec:Data_and_Methods}

This section briefly introduces the geophysical data and meteorological forcings we use to feed the data-driven models (\cref{sec:Data_and_Methods:data}), as well as our approach to train the models (\cref{sec:Data_and_Methods:training_methods}) in order to answer the two research questions (for details, see \cref{sec:Data_and_Methods:training_methods:Input_sequence,sec:Data_and_Methods:training_methods:period_length_basins}, respectively).

\subsection{Data}
\label{sec:Data_and_Methods:data}
In our experiments, we use the CAMELS dataset, as it is commonly used as sufficiently large and diverse to support some degree of generalization and application to other scenarios~\citep{Newman2017Benchmark, Addor2018Ranking, Kratzert2019Benchmark, Mizukami2019Metrics}.

We replicate the setup in \citet{Kratzert2019Benchmark}\ and train and test our models using the same data from the CAMELS dataset~\citep{Newman2014Camels, Addor2017CAMELS}.
This dataset, curated by the US National Center for Atmospheric Research, contains daily streamflow measurements for 671~basins across the continental United States.
Following \citet{Newman2017Benchmark} and \citet{Kratzert2019Benchmark}, we only use 531~basins and discard basins that exhibit large discrepancies in their areas as calculated using different strategies.
The dataset additionally provides three sets of daily basin-averaged meteorological forcings for each basin.
Again, we follow \citet{Kratzert2019Benchmark} and use the Maurer forcings~\citep{Maurer2002Forcings} that include daily cumulative precipitation, minimum and maximum air temperature, average short-wave radiation, and vapor pressure, spatially aggregated for each basin. 
Further, we use 27 of the static basin characteristics distributed with the CAMELS dataset~\citep{Addor2017CAMELS}; refer to \cref{app:static} for a list of these characteristics.
The spatial aggregation is derived from an original, gridded dataset with a resolution of \SI{1/8}{\degree}.

\subsection{Training and Evaluation Procedures}
\label{sec:Data_and_Methods:training_methods}
Building upon the open-source code of \citet{Kratzert2019Benchmark}, we are able to largely replicate their Entity-Aware LSTM (EA-LSTM) results with differences in median Nash--Sutcliffe-efficiency (NSE) well below $0.01$, which can be easily attributed to non-determinism in the training process.
We use the basin-averaged NSE from \citet{Kratzert2019Benchmark} as the loss function for all models.
\cref{app:setup} provides details about the tuning and training procedures, as well as our computational environment.

As we train our models, we consider the following three dimensions of training dataset size:
\begin{description}
    \item[Training period length.]
        The overall available amount of historical data used to train on a given basin. This is perhaps the most obvious and most commonly exploited dimension in streamflow predictions.
    \item[Hydro-geo-climatic diversity.]
        The number of basins that we use to train a single model.
        While the advantage of greater geographic, hydrologic, and geophysical diversity may at first seem counter-intuitive, previous work shows that models are able to generalize knowledge across basins \citep{Kratzert2018LSTM}.
        For instance, a model may have never seen a period of drought in a certain basin during training, but it may have seen a drought in a basin with similar characteristics.
        In such a case, an ideal model would transfer the knowledge from the second basin to the first and yield accurate predictions, even though it was never trained to predict droughts on the first basin.
        Instead of merely comparing the performance impact between training on a single basin and training on the full available basin set, we study this effect on various degrees of diversity and provide a direct quantitative comparison.
        In our experiments, we achieve different degrees of diversity by training and testing on different amounts of randomly selected basins.
        We refrain from considering hydrologic measures of similarity, since there is no commonly agreed-upon similarity metric, and any choice would neglect certain geographic, hydrologic, or geophysical aspects.
    \item[Input sequence length.] 
        Training period length and hydro-geo-climatic diversity determine the overall number $N$ of observations that we show the models during training.
        The input sequence length $k$ is a fundamentally different dimension, as it varies the amount of data we feed a model to predict an individual sample: when predicting streamflow $q_t$ for day $t \in [1, \dots, T]$, we feed the model forcings from the previous $k$ days $x_{t-k+1}, \dots, x_t$.
        Hence, we can vary $k$ to change the amount of data we feed the models to fit an individual streamflow observation $q_t$.
        Here, $k$ corresponds to the length of the forcing sequence that leads up to an individual sample.
        Note that, unlike the other two dimensions, the input sequence length is independent of the overall dataset size (apart from the training period length being an upper bound).
        Rather, it is a tunable parameter of model training; for instance, too short sequences (small $k$) might not contain sufficient information for accurate predictions, and too long sequences (large $k$) might make it challenging for a model to extract the most important patterns.
\end{description} 
A priori, it is not obvious whether predictions improve with additional training data in all or some of these dimensions.
Moreover, it is likely that a model's response to these dimensions is not independent:
If the training period is limited, it may for instance be beneficial to reduce the input sequence length, too, since the models will not see enough training samples to infer long-term relationships between input time series and streamflow response.

To estimate the training set size dimensions' effects on prediction quality, we compare the accuracy of XGBoost and EA-LSTMs when trained on differently-sized subsets of the CAMELS dataset.
This method is called split-sample testing in hydrology.
We individually restrict the number of available training years, the number of basins, and the input sequence length to varying degrees.

All models obtain as input 27 static catchment attributes from the CAMELS dataset~\citep{Addor2017CAMELS} that provide properties of climate, vegetation, soil, and topography (cf.\@~\cref{app:static}), as well as the five normalized meteorological variables from the Maurer forcings~\citep{Maurer2002Forcings} provided with the CAMELS dataset.
Due to limited computational resources, we focus on the EA-LSTM architecture as introduced by \citet{Kratzert2019Benchmark} and do not evaluate standard LSTMs.
As both architectures are very similar, we do, hovewer, believe that our results largely apply to standard LSTMs, as well.

\subsubsection{Input Sequence Length}
\label{sec:Data_and_Methods:training_methods:Input_sequence}
We use input sequence lengths $k$ of $10$, $30$, $100$, $270$, and $365$ previous days' forcings for EA-LSTMs.
For XGBoost, we feed the sequences of length $k$ as flattened vectors of $5k$ variables, concatenated with the $27$ catchment attributes to a vector of length $5k + 27$.
We only use input sequences of length $10$, $30$, and $100$, but no longer ones, as the predictions with $100$ days are already worse than the predictions using only $30$ days.
Likely, this is due to the fact that the sequence length affects the input dimension by a multiplicative factor, as we use a flattened vector as input.
An input sequence of 100 days therefore already leads to a $100 \times 5 + 27 = 527$-dimensional input space, where training is challenging.
Longer sequences also drastically increase the runtime of the already computationally intensive hyperparameter search.

\subsubsection{Training Period Length and Number of Basins}
\label{sec:Data_and_Methods:training_methods:period_length_basins}
Besides the input sequence length, we evaluate the quality of each model by varying the amount of training data in terms of the number of training years and the number of basins.

All training periods start from October 1999 and last three, six, or nine years.
The basin sets that emulate hydro-geo-climatic diversity are random subsets of $13$ ($2.5\%$), $26$ ($5\%$), $53$ ($10\%$), $265$ ($50\%$), and $531$ ($100\%$) basins. 
For each setup with a basin set size below $531$, we evaluate five random basin selections of that size.
To ensure comparability, setups with the same basin set size use the exact same five basin sets.
We do not conduct experiments on single basins, although this may at first seem like a suitable benchmark.
The reason for this is that previous shows that there is never a reason to train a single-basin model just by itself. Instead, one should either train on a larger corpus that includes this basin, or transfer a multi-basin model to the individual basin \citep{Kratzert2019Benchmark, Kratzert2018LSTM}.
While a large-scale evaluation of transfer learning strategies would be a valuable contribution, we leave this to future work.

For reference, we compare the models' results with the accuracy of mHM, which is one of the two best traditional hydrologic models in the benchmark by \citet{Kratzert2019Benchmark}.
Note that unlike XGBoost and EA-LSTM, mHM is calibrated on all nine years of data for each basin individually, which makes it a more demanding benchmark than, for instance, a CONUS-wide calibrated model.
We do not run mHM ourselves, but use the predictions from \citet{Mizukami2019Metrics}.
Following exactly the same setup as \citet{Kratzert2019Benchmark}, we test all models on the test period from October 1989 until September 1999, and we evaluate them on the respective set of basins that they were trained on (in hydrologic terms, we perform temporal validation).
To reduce errors due to random initialization, we train each XGBoost and EA-LSTM model with eight different seeds and evaluate the ensemble of their averaged predictions.

For each of the 15 combinations of three training period lengths and five basin set sizes, we select the EA-LSTM and XGBoost model trained with the input sequence length $k$ that results in the best median NSE across all basins and across the five random basin sets.
As a result, we obtain two distributions $F_{\text{EA-LSTM}}$ and $F_{\text{XGBoost}}$ of NSE values for the models.
Each distribution has a sample size of five times the number of basins (except for the case of all $531$ basins, where we do not have five random basin sets).
We use a Kolmogorov--Smirnov significance test to test the null hypothesis that the distributions of NSE values are identical.
For this, we use Python's scipy function \texttt{scipy.stats.ks\_2samp}.
This test is based on the maximum absolute difference between $F_{\text{EA-LSTM}}$ and $F_{\text{XGBoost}}$.
The significance test's $p$-value denotes the probability that the NSE values are at least as different as in our experiment even though they come from the same distribution.
To account for the large number of 15 significance tests, we apply Bonferroni correction~\citep{Mittelhammer2000Econometric} and test our hypotheses at $\alpha = 0.01/15$.
Hence, we accept the hypothesis of identical distributions for $p$-values below $0.01/15$ and reject it otherwise.
Further, we estimate the corresponding effect size as Cohen's $d$~\citep{Cohen2013statistical}.
This effect size is a metric for the difference of distributions; it measures the difference between the means of the two NSE distributions, normalized by their combined standard deviation ($d = (\mu_{\text{EA-LSTM}} - \mu_{\text{XGBoost}})/\sigma_{(\text{EA-LSTM}, \text{XGBoost})}$).
The larger $d$, the further apart are the distribution means.

While the effect size $d$ measures the difference between the two models' NSE distributions, we quantify the quality of the models' predictions as the area $A$ under the curve of their cumulative NSE distributions.
Ideally, this value is zero, with an NSE of one on all basins.
As models yield smaller NSEs, the distribution shifts leftward and $A$ increases.
Larger values of $A$ therefore correspond to worse overall performance.

\section{Results}
\label{sec:Results}

In the following, we present the results of our experiments to address our two research questions (\cref{sec:Results:Input_sequence,sec:Results:period_length_basins}, respectively).

\subsection{Input Sequence Length}
\label{sec:Results:Input_sequence}
\begin{table}[pos=ht]
    \centering
    \renewcommand{\arraystretch}{1.2}
    \caption{Input sequence lengths $k$ for XGBoost and EA-LSTM that yield the best median NSE for each number of basins (rows) and training years (columns) (Section~(A)).
    Section~(B) shows the difference in median NSE between the best and next-shorter sequence length.
    For the reader's convenience, we color-code the cell entries; lighter colors correspond to larger values.}
    \label{tab:seqlens}
    \begin{tabular}{ccrrccccccc}
        \toprule
         & & & & \multicolumn{7}{c}{Training years} \\
         & & &  & \multicolumn{3}{c}{XGBoost} & & \multicolumn{3}{c}{EA-LSTM} \\
        \cmidrule(lr){5-7}\cmidrule{9-11}
         & & & & 3 & 6 & 9 & & 3 & 6 & 9\\
        \midrule
        \multirow{5}{*}{\textbf{(A)}} & \multirow{5}{*}{\rotatebox[origin=c]{90}{\parbox[c]{2cm}{\centering Number (percentage) of basins}}} & 13 & (2.5\%) & \seqten & \seqten & \seqten & & \seqten & \seqten & \seqthirty\\
         & & 26 & (5\%) & \seqten & \seqthirty & \seqthirty & & \seqten & \seqthirty & \seqthirty \\
         & & 53 & (10\%) & \seqten & \seqthirty & \seqthirty & & \seqthirty & \seqthirty & \seqhundred \\
         & & 265 & (50\%) & \seqthirty & \seqthirty & \seqthirty & & \seqhundred & \seqtwoseventy & \seqyear \\
         & & 531 & (100\%) & \seqthirty & \seqthirty & \seqthirty & & \seqtwoseventy & \seqyear & \seqyear \\

        \midrule

        \multirow{5}{*}{\textbf{(B)}} & \multirow{5}{*}{\rotatebox[origin=c]{90}{\parbox[c]{2cm}{\centering Number (percentage) of basins}}} & 13 & (2.5\%) & -- & -- & -- & & -- & -- & \cellcolor[HTML]{ef7c51}\textcolor[HTML]{000000}{$0.046$} \\ 
        & & 26 & (5\%) & -- & \cellcolor[HTML]{8104a7}\textcolor[HTML]{f1f1f1}{$0.022$} & \cellcolor[HTML]{d04d73}\textcolor[HTML]{000000}{$0.044$} & & -- & \cellcolor[HTML]{f3854b}\textcolor[HTML]{000000}{$0.048$} & \cellcolor[HTML]{febb2b}\textcolor[HTML]{000000}{$0.058$} \\ 
        & & 53 & (10\%) & -- & \cellcolor[HTML]{ae2892}\textcolor[HTML]{f1f1f1}{$0.033$} & \cellcolor[HTML]{d45270}\textcolor[HTML]{000000}{$0.045$} & & \cellcolor[HTML]{e97257}\textcolor[HTML]{000000}{$0.044$} & \cellcolor[HTML]{fada24}\textcolor[HTML]{000000}{$0.063$} & \cellcolor[HTML]{6c00a8}\textcolor[HTML]{f1f1f1}{$0.014$} \\
        & & 265 & (50\%) & \cellcolor[HTML]{e76e5b}\textcolor[HTML]{000000}{$0.053$} & \cellcolor[HTML]{f79044}\textcolor[HTML]{000000}{$0.062$} & \cellcolor[HTML]{fdb42f}\textcolor[HTML]{000000}{$0.070$} & & \cellcolor[HTML]{7e03a8}\textcolor[HTML]{f1f1f1}{$0.017$} & \cellcolor[HTML]{350498}\textcolor[HTML]{f1f1f1}{$0.005$} & \cellcolor[HTML]{20068f}\textcolor[HTML]{f1f1f1}{$0.002$} \\
        & & 531 & (100\%) & \cellcolor[HTML]{f58c46}\textcolor[HTML]{000000}{$0.061$} & \cellcolor[HTML]{febd2a}\textcolor[HTML]{000000}{$0.072$} & \cellcolor[HTML]{fada24}\textcolor[HTML]{000000}{$0.078$} & & \cellcolor[HTML]{43039e}\textcolor[HTML]{f1f1f1}{$0.007$} & \cellcolor[HTML]{350498}\textcolor[HTML]{f1f1f1}{$0.005$} & \cellcolor[HTML]{16078a}\textcolor[HTML]{f1f1f1}{$0.001$} \\
        \bottomrule
    \end{tabular}
    \renewcommand{\arraystretch}{1}
\end{table}
\cref{tab:seqlens}A shows the sequence lengths that result in the best median NSE across all basins and across the five random basin sets for a given training period length and number of basins.
As expected, we can generally say that optimal sequence lengths increase with the amounts of spatial and temporal observations.
For XGBoost, sequences of length 30 consistently yield better results than those of length 100, which confirms our expectation that the increasing training space complexity outweighs the added information longer sequences provide.
EA-LSTMs do not exhibit this behavior, as we find that the best sequence length increases beyond 30.
On the three largest training set configurations it even reaches 365, which is the longest sequence length we consider.

\cref{tab:seqlens}B shows the difference between the median NSE on the best and the next-shorter sequence length at each training set size.
While this difference increases with training set size for XGBoost (likely due to the constant sequence length of $30$ on the larger training sets), it decreases for EA-LSTMs.

\subsection{Training Period Length and Number of Basins}
\label{sec:Results:period_length_basins}

The following analysis describes our results with respect to our second research question, where we examine how training period length and the number of basins in the training dataset affect the prediction quality of XGBoost and EA-LSTM.
In Section~\ref{sec:Results:period_length_basins:distributions}, we focus on the two models' NSE distributions on the differently-sized training sets.
Section~\ref{sec:Results:period_length_basins:medians} analyzes the models' correlation of median accuracy across all basins and training set size as well as spatial NSE patterns.

\subsubsection{Analysis of Prediction Accuracy Distributions}
\label{sec:Results:period_length_basins:distributions}

\begin{figure}[pos=t]
    \centering
    \noindent\includegraphics[width=0.75\textwidth]{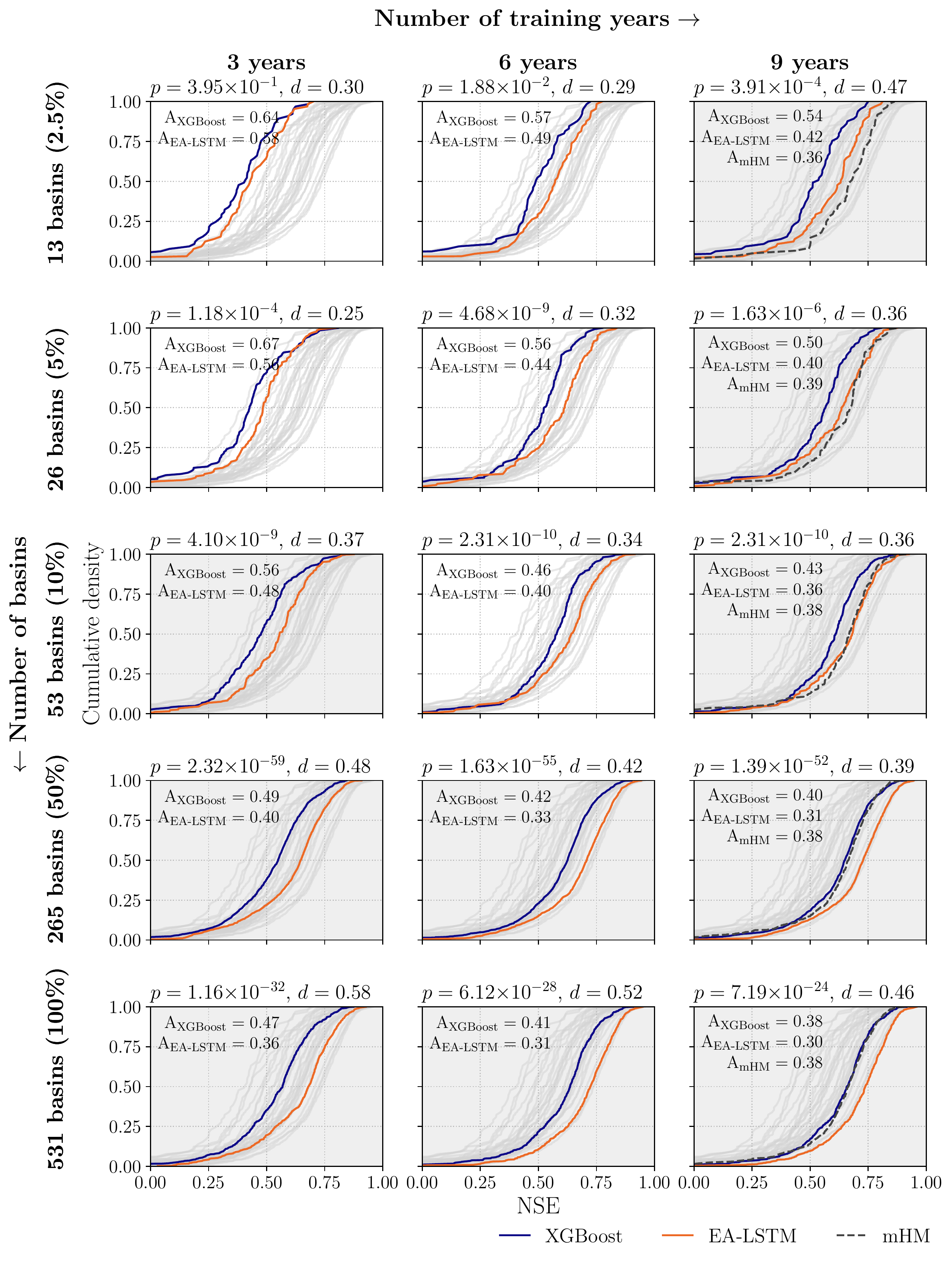}
    \caption{Empirical cumulative NSE distributions for XGBoost (blue) and EA-LSTM (orange) at varying amounts of training data in terms of training period (columns) and number of basins (rows) using the optimal input sequence length $k$ (cf.\@~\cref{tab:seqlens}A).
    For reference, the dashed gray lines show the distribution for the mHM model trained on nine years for each basin individually.
    Each plot shows the $p$-value of a Kolmogorov--Smirnov significance test and the effect size as Cohen's $d$.
    Plots with gray backgrounds correspond to combinations with $p < 0.01/15$ and $d > 0.35$, which indicates distribution pairs with at least small to medium differences.
    $A_{\{\text{XGBoost, EA-LSTM, mHM}\}}$ indicate the area under the distribution curves; lower values are better.}
    \label{fig:gridplot}
\end{figure}

\cref{fig:gridplot} and \cref{tab:results} provide an overview of our experimental results regarding the influence of training period length and number of basins used for training on the prediction accuracy in the test period.

Each plot in a particular row and column in \cref{fig:gridplot} corresponds to one of the 15 combinations of basin set size and training period length, and it shows the empirical cumulative NSE distributions for XGBoost and EA-LSTM when we use the respectively best input sequence length (cf.\@~\cref{tab:seqlens}A).
For comparison, the dashed line shows the cumulative NSE distribution for the basin-wise calibrated mHM model trained on nine years of data.
An ideal model with perfect predictions would exhibit a ``\reflectbox{\textsf{L}}''-shaped distribution, yielding high NSE values across all basins; the corresponding area $A$ under the curve for such a model would be zero.

\cref{fig:gridplot} specifically highlights cases in gray where the Kolmogorov--Smirnov significance test supports rejecting the null hypothesis of identical distributions for XGBoost and EA-LSTM at $p < 0.01/15$ (after Bonferroni correction) and Cohen's $d > 0.35$.
The threshold of $d > 0.35$ is halfway between the scales suggested by \citet{Sawilowsky2009Effect} for ``small'' and ``medium'' effect sizes. 

As expected, more training data---both in terms of training period length and number of basins---increases the prediction accuracy of all models.
Visually, this is shown by the rightward shift of the distributions in \cref{fig:gridplot}.
Quantitatively, both models' area under the distribution curve decreases, from $A_{\text{XGBoost}} = 0.64$, $A_{\text{EA-LSTM}} = 0.58$ on the smallest training set to $A_{\text{XGBoost}} = 0.38$ and $A_{\text{EA-LSTM}} = 0.30$ on the largest configuration.
Compared to the mHM benchmark (which is trained on nine years and each basin individually), XGBoost's accuracy is roughly on par in the largest configuration.
The EA-LSTM outperforms mHM already on $53$ basins and gains further advantage with even larger basin sets.
In cases where both training period and number of basins are very limited (for instance, the cases of three to six years and 13 basins), XGBoost and EA-LSTM result in NSE distributions that do not differ significantly (large $p$-values), although the EA-LSTM has a slightly better median NSE.
With access to more training data, EA-LSTMs begin to significantly outperform XGBoost at small ($d > 0.2$) to medium ($d > 0.5$) effect sizes~\citep{Sawilowsky2009Effect}.
The effect size $d$ grows with the number of basins, but has no clear relationship to the number of training years, whereas the models' area $A$ under the distribution curve decreases with both the number of basins and training years.

Following the Kolmogorov--Smirnov significance test, we can reject the hypothesis of identical NSE distributions for the conditions of three to six years and 26 or more basins, and all conditions with nine years at $p < 0.01/15$.

\cref{tab:results} lists the models' minimum, median, and maximum NSE scores for each of the 15 training set configurations, as well as the average percentage of ``failures'' (basins with an NSE below zero) across the five different random basin selections.
All numbers refer to the models using the optimal input sequence lengths (cf.\@~\cref{tab:seqlens}A).
The best value for each metric is highlighted in bold font.
Comparing the extremes in \cref{tab:results}, we further note that, with few exceptions, the EA-LSTM model results in higher minimum and maximum NSE values as well as a smaller percentage of failed basins ($\text{NSE} \leq 0$).
Notably, unlike median and maximum, both models' minimum NSE values neither clearly improve with training period length nor with the number of basins.
The same holds true for the average percentage of failed basins.
This is especially surprising for the training period length, as the models obtain more observations for the same set of basins, which should improve their predictions.
Explanations for this might be that the added samples---most of which show modest streamflow---entice the models to more conservative predictions near the bulk of the training samples, which results in worse NSE values when a more extreme event occurs, or erroneous streamflow observations.
\begin{table}[pos=t]
\robustify\bfseries
\caption{Minimum, median, and maximum NSE scores and average percentage of failed basins ($p_{\text{failed}} [\%]$, $\text{NSE} \leq 0$) on the test (validation) period (Oct.\ 1989 to Sep.\ 1999) for XGBoost and EA-LSTM models, trained with different amounts of training years ($N_{\text{years}}$) and basins ($N_{\text{basins}}$). The values are calculated across five different random basin selections (minimum is aggregated as minimum, median as median, maximum as maximum, and $p_{\text{failed}}$ as the average percentage of failed basins in each random basin set).
In each row, the best values for each metric are highlighted in bold.}
\label{tab:results}
\centering
\begin{tabularx}{0.8\textwidth}{RRS[table-format=1.2, detect-weight]S[table-format=1.2, detect-weight]S[table-format=1.2, detect-weight]S[table-format=1.2, detect-weight]S[table-format=1.2, detect-weight]S[table-format=1.2, detect-weight]S[table-format=1.2, detect-weight]S[table-format=1.2, detect-weight]}
\toprule
 \multirow{2}{*}{$N_{\text{years}}$} & \multirow{2}{*}{$N_{\text{basins}}$} & \multicolumn{4}{c}{XGBoost} & \multicolumn{4}{c}{EA-LSTM} \\
 \cmidrule(r){3-6}\cmidrule(r){7-10}
 &  & {Min} & {Median} & {Max} & {$p_{\text{failed}} [\%]$} & {Min} & {Median} & {Max} & {$p_{\text{failed}} [\%]$}\\
\midrule
3 & 13 & -0.64 & 0.41 & 0.69 & 4.62 & \bfseries -0.31 & \bfseries 0.43 & \bfseries 0.70 & \bfseries 1.54 \\
 & 26 & -4.71 & 0.43 & \bfseries 0.81 & 4.62 & \bfseries -1.58 & \bfseries 0.48 & 0.80 & \bfseries 3.08 \\
 & 53 & -1.98 & 0.48 & 0.85 & 2.26 & \bfseries -1.36 & \bfseries 0.56 & \bfseries 0.88 & \bfseries 0.38 \\
 & 265 & -1.92 & 0.55 & 0.84 & 1.81 & \bfseries -0.25 & \bfseries 0.65 & \bfseries 0.91 & \bfseries 0.30 \\
 & 531 & -1.34 & 0.57 & 0.87 & 1.51 & \bfseries 0.03 & \bfseries 0.68 & \bfseries 0.93 & \bfseries 0.00 \\
\midrule
6 & 13 & \bfseries -1.06 & 0.49 & 0.72 & 4.62 & -1.65 & \bfseries 0.57 & \bfseries 0.77 & \bfseries 1.54 \\
 & 26 & -4.14 & 0.53 & 0.78 & 3.08 & \bfseries 0.01 & \bfseries 0.61 & \bfseries 0.84 & \bfseries 0.00 \\
 & 53 & -1.15 & 0.58 & 0.86 & 0.75 & \bfseries -0.26 & \bfseries 0.64 & \bfseries 0.91 & \bfseries 0.38 \\
 & 265 & -3.08 & 0.63 & 0.90 & 1.36 & \bfseries -0.20 & \bfseries 0.71 & \bfseries 0.94 & \bfseries 0.45 \\
 & 531 & -1.75 & 0.64 & 0.91 & 0.94 & \bfseries -0.90 & \bfseries 0.72 & \bfseries 0.95 & \bfseries 0.56 \\
\midrule
9 & 13 & -1.39 & 0.54 & 0.75 & 3.08 & \bfseries -0.16 & \bfseries 0.63 & \bfseries 0.81 & \bfseries 1.54 \\
 & 26 & -2.41 & 0.57 & 0.81 & 2.31 & \bfseries -0.01 & \bfseries 0.64 & \bfseries 0.87 & \bfseries 0.77 \\
 & 53 & -1.22 & 0.61 & 0.89 & 1.13 & \bfseries -0.02 & \bfseries 0.68 & \bfseries 0.93 & \bfseries 0.38 \\
 & 265 & -4.91 & 0.65 & 0.91 & 1.43 & \bfseries -0.96 & \bfseries 0.73 & \bfseries 0.95 & \bfseries 0.23 \\
 & 531 & \bfseries -1.21 & 0.66 & 0.91 & 1.13 & -1.38 & \bfseries 0.74 & \bfseries 0.96 & \bfseries 0.19 \\
\bottomrule
\end{tabularx}
\end{table}

\subsubsection{Analysis of Median Prediction Accuracy}
\label{sec:Results:period_length_basins:medians}
\begin{figure}[pos=ht]
    \centering
    \noindent\includegraphics[width=\textwidth]{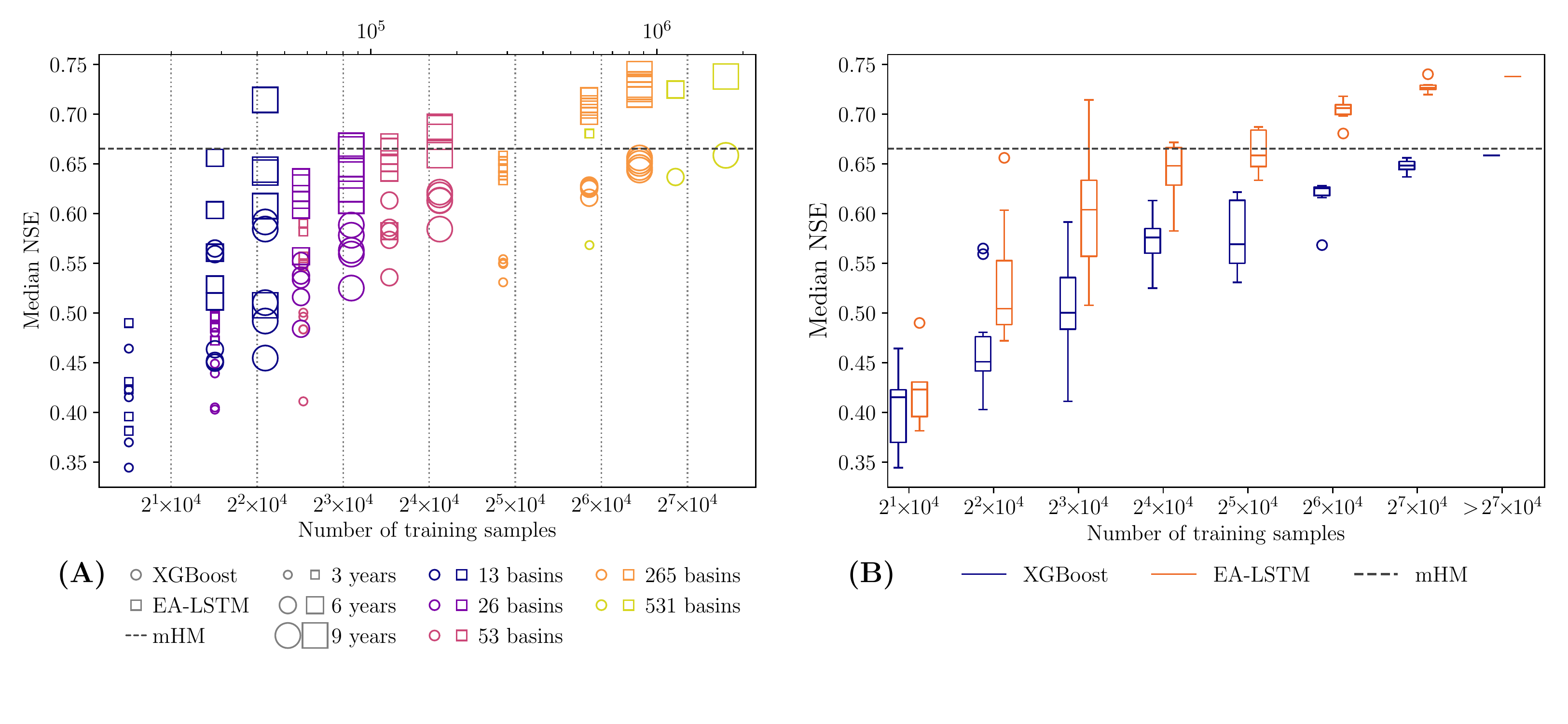}
    \caption{Relationship between number of training samples ($N_{\text{basins}} \times (N_{\text{years}} \times 365)$) and median NSE across all basins.
    Markers in panel~(A) represent the median NSE for XGBoost (circles) and EA-LSTM (squares) on a certain training period length (three to nine years, marker size), number of basins (13--531, marker color). 
    As we use five random basin subsets of each size, we report five median NSE scores for the combinations with less than 531 basins.
    The dashed horizontal line indicates the median NSE of the mHM benchmark trained on nine years of each basin individually.
    Panel~(B) groups these median NSE values for XGBoost (blue) and EA-LSTM (orange) into the orders of magnitudes indicated by the gray lines in panel~(A).
    The boxplots aggregate each model's results for combinations with similar amounts of training data.
    The boxes extend from lower $p_{25}$ to upper $p_{75}$ quartile and have a line at the median.
    The whiskers reach to the last data point that is up to $1.5 \times (p_{75} - p_{25})$ beyond the boxes' ends, circles indicate outliers beyond those points.
    Note the logarithmic scale of the x-axes.}
    \label{fig:samples-vs-nse}
\end{figure}

\cref{fig:samples-vs-nse} shows the relation between the number of training samples and the median NSE scores across all basins for XGBoost and EA-LSTM.
\cref{fig:samples-vs-nse}A denotes XGBoost as circles and EA-LSTM as squares.
It distinguishes training set size between training period length (marker size) and number of basins (marker color).
The fact that squares are almost always above circles of the same size and color indicates again that EA-LSTMs outperform XGBoost in most cases.
We note that both models benefit not only from longer training periods, but also from larger basin subsets:\
The models' predictions improve even if the number of training years remains constant but only the number of basins increases (lighter colors of same size in \cref{fig:samples-vs-nse}A).
Although there are not many examples, it appears that a larger number of training years is more effective than a larger number of basins at the same training set size: darker, larger shapes are mostly above lighter, smaller ones at the same x-axis value in \cref{fig:samples-vs-nse}A.

In a more aggregated view, \cref{fig:samples-vs-nse}B plots the models' median NSE distributions bucketed into the orders of training set size magnitude denoted by the vertical lines in \cref{fig:samples-vs-nse}A.
As already shown in \cref{fig:gridplot}, we affirm three things: both models start at similar NSE ranges, EA-LSTMs overtake XGBoost as the amount of training samples increases, and more data generally result in better prediction quality.
There is a strong trend of increasing median NSE scores with larger training set sizes up to about $2^4 \times 10^4$ samples.
For even larger training sets, the median NSE values mostly continue to grow, but we clearly see smaller improvements.  

\begin{figure}[pos=p]
   \centering
   \noindent\includegraphics[width=0.9\textwidth]{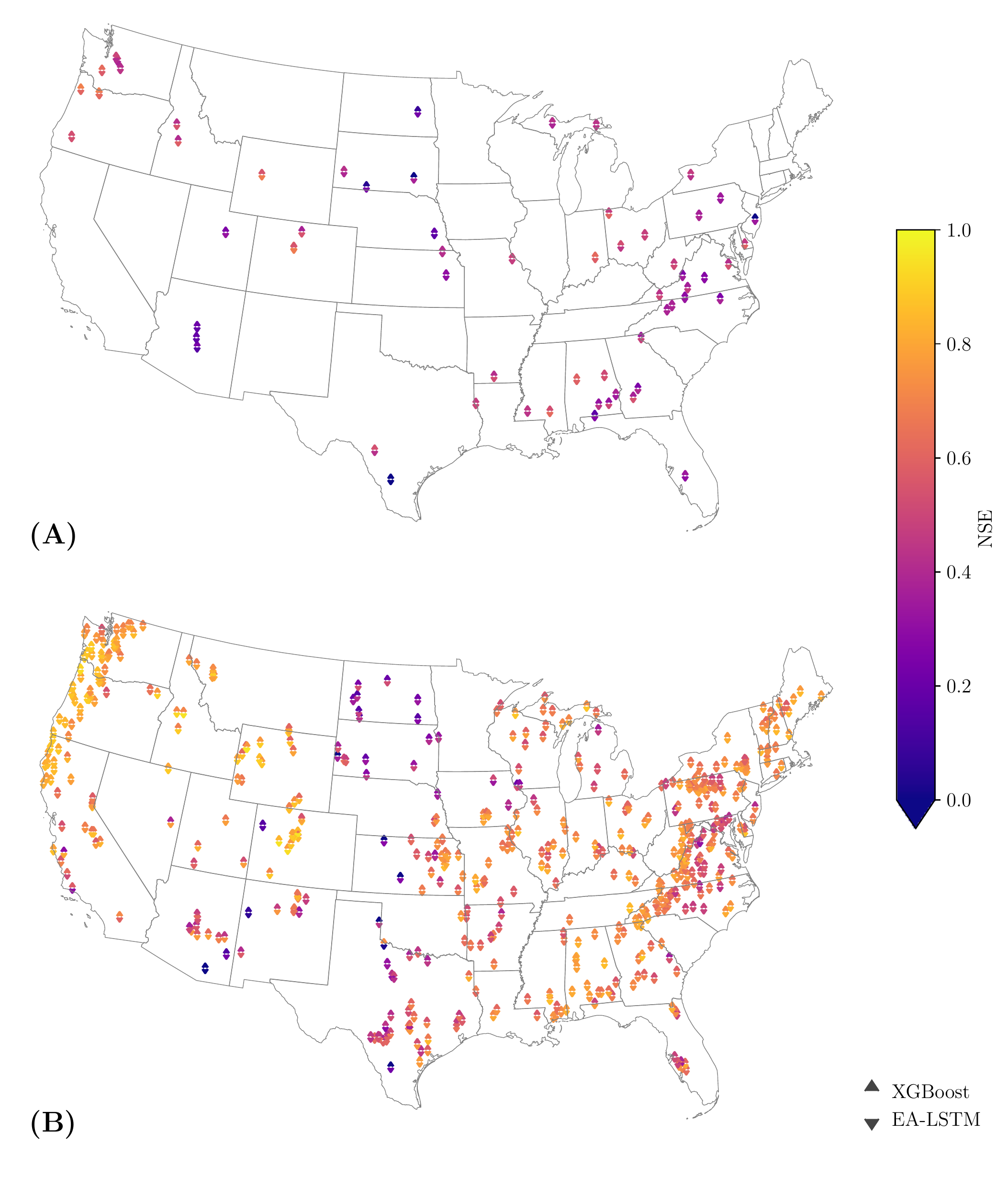}
   \caption{Heatmap visualization of the XGBoost and EA-LSTM models' NSE values for each basin.
   Panel~(A) shows the NSEs on the smallest configuration of three years and 13 basins.
   It displays basins that are part of at least one random 13-basin subset.
   Where a basin is part of multiple subsets, we show the average NSE.
   Panel~(B) shows the NSEs on the largest configuration of nine years and all 531 basins.}
   \label{fig:nsemap}
\end{figure}

As the heatmaps in \cref{fig:nsemap} show, the accuracy of XGBoost and EA-LSTMs on individual basins is strongly correlated for both small (Pearson correlation coefficient $0.84$, \cref{fig:nsemap}A) and large datasets (Pearson correlation coefficient $0.78$, \cref{fig:nsemap}B).
Furthermore, in both figures it appears that both models make poor predictions on the same basins, with the exception of two basins in \cref{fig:nsemap}B; one where XGBoost yields much worse predictions than the EA-LSTM model (NSEs $0.29$ vs.\ $0.81$), and one vice versa (NSEs $0.73$ vs.\ $0.02$).
Based on manual examination of the results, the difference in these striking instances can largely be explained by few days in which one model vastly overestimates the daily streamflow.
As the NSE metric is highly sensitive to such outliers, this decreases the affected basins' overall score.

\section{Discussion and Future Work}
\label{sec:Discussion}

\subsection{Input Sequence Length}
Our results outline dependencies between the analyzed dimensions, as larger training sets in terms of training period length and number of basins call for longer input sequences.
The comparison to our findings in the initial version of this study~\citep{Gauch2019Feeding}---where we use fixed sequence lengths for both models---shows the importance of carefully chosen input sequence lengths: with an adaptive sequence length, the EA-LSTM's accuracy becomes comparable to XGBoost's on small datasets.   
On smaller datasets, we therefore recommend careful tuning of the input sequence length for EA-LSTMs.

Considering the small difference in accuracy between sequence lengths of 365 and 270 for EA-LSTM (cf.\@~\cref{tab:seqlens}B), it is likely that sequence lengths beyond 365 do not further improve predictions, and that the optimal sequence length stabilizes for training set sizes beyond the ones we explore in this study.
In contrast to the EA-LSTM, XGBoost exhibits a smaller and much more robust optimal sequence length of $k=30$ except for very small training sets.
To us, it seems that XGBoost's optimal sequence length would only increase beyond $k=30$ for substantially larger training sets than the ones examined here.

\subsection{Training Period Length and Number of Basins}
While---somewhat unsurprisingly---more training data generally improve predictions, our findings also paint a more nuanced picture:\
We note that not only longer training periods, but also data from additional basins increase the models' prediction accuracy.
This is an important finding for two reasons:
First, the advent of machine learning in hydrology gave rise to an abundance of studies that train data-driven models on just one or very few basins (e.g., \citet{Hu2018DL}, \citet{Dastorani2018ML}).
Although even such selected-basin architectures might outperform traditional models and be helpful in operational scenarios, our results show that leveraging large amounts of basins yields better models, eliminates the need to train individually on each basin, \textit{and} opens perspectives towards spatial generalization.
Second, it suggests that the models actually infer relations between catchment characteristics and streamflow patterns, rather than merely overfitting on the given set of basins in the training data.
Our study corroborates the results from \citet{Kratzert2018LSTM}, where the authors show that LSTMs should be trained on multiple basins. 
Building upon these results, we empirically show that additional basins continue to improve the models' performance even if there are already hundreds of basins in the training set.
In other words, it is not enough to train models on more than one basin, but one should strive to use as many basins as possible.
While this holds for both models, the growing effect size $d$ and the decreasing area $A$ under the cumulative distribution curve indicate that EA-LSTMs benefit more than XGBoost from larger basin sets.
Our study focuses on NSE values, which emphasize accuracy on high-flow events.
To analyze the models with regards to their accuracy on low flows, we further calculated the NSE of the logarithmic discharge (logNSE, results not shown).
The logNSE also generally improves with increasing training set size, although not quite as clearly as the NSE, likely because the models were not trained to minimize logNSE.

The ability to generalize knowledge across basins is especially relevant for predictions in data-poor regions, as it implies that we might improve results in these areas by leveraging additional training data from similar basins in other, data-rich regions. This is one step towards employing machine learning approaches to predictions in ungauged basins.

Based on our experiments, we recommend LSTMs over the tree-based XGBoost for streamflow prediction. 
Although LSTMs require more careful tuning of the input sequence length $k$, they benefit from their ability to directly ingest time-series data and outperform XGBoost significantly in almost all our experiments.

\subsection{Future Work}
In future work, it appears worthwhile to explore the potential of transfer learning for data-poor basins, as we can pre-train models on areas where there are copious amounts of data and then subsequently fine-tune them using the scarce data available for the target basin.
This in particular seems like a promising direction for applying data-driven models to predictions on ungauged basins.
\clearpage
\appendix
\section{Static Basin Attributes}
\label{app:static}
\cref{app:tab:static} lists the 27 static basin attributes we use to train our models in this study.
They are the same attributes as used by \citet{Kratzert2019Benchmark}.
\begin{table}[pos=ht]
\centering
\caption{Static basin attributes from the CAMELS dataset used in this study. (Source: \citet{Addor2017CAMELS})}
\label{app:tab:static}
\begin{tabularx}{\textwidth}{lX}
	\toprule
	Variable & Description\\
	\midrule
	\texttt{p\_mean} & Mean daily precipitation.\\
	\texttt{pet\_mean} & Mean daily potential evapotranspiration.\\
	\texttt{aridity} & Ratio of mean PET to mean precipitation.\\
	\texttt{p\_seasonality} & Seasonality and timing of precipitation. Estimated by representing annual precipitation and temperature as sin waves. Positive (negative) values indicate precipitation peaks during the summer (winter). Values of approx.\ 0 indicate uniform precipitation throughout the year.\\
	\texttt{frac\_snow\_daily} & Fraction of precip.\ falling on days with temperatures $< \SI{0}{\celsius}$.\\
	\texttt{high\_prec\_freq} & Frequency of high precipitation days ($\leq 5 \times \texttt{p\_mean}$).\\
	\texttt{high\_prec\_dur} & Average duration of high precipitation events (number of consecutive days with $\leq 5 \times \texttt{p\_mean}$).\\
	\texttt{low\_prec\_freq} & Frequency of dry days ($<$ 1 mm/day).\\
	\texttt{low\_prec\_dur} & Average duration of dry periods (number of consecutive days with precipitation $<$ 1 mm/day).\\
	\texttt{elev\_mean} & Catchment mean elevation.\\
	\texttt{slope\_mean} & Catchment mean slope.\\
	\texttt{area\_gages2} & Catchment area.\\
	\texttt{forest\_frac} & Forest fraction.\\
	\texttt{lai\_max} & Maximum monthly mean of leaf area index (LAI).\\
	\texttt{lai\_diff} & Difference between the max.\ and min.\ mean of the LAI.\\
	\texttt{gvf\_max} & Maximum monthly mean of green vegetation fraction (GVF).\\
	\texttt{gvf\_diff} & Difference between the max.\ and min.\ monthly mean GVF.\\
	\texttt{soil\_depth\_pelletier} & Depth to bedrock (maximum \SI{50}{\metre}).\\
	\texttt{soil\_depth\_statsgo} & Soil depth (maximum \SI{1.5}{\metre}).\\
	\texttt{soil\_porosity} & Volumetric porosity.\\
	\texttt{soil\_conductivity} & Saturated hydraulic conductivity.\\
	\texttt{max\_water\_content} & Maximum water content of the soil.\\
	\texttt{sand\_frac} & Fraction of sand in the soil.\\
	\texttt{silt\_frac} & Fraction of silt in the soil.\\
	\texttt{clay\_frac} & Fraction of clay in the soil.\\
	\texttt{carb\_rocks\_frac} & Fraction of the catchment area characterized as ``Carbonate sedimentary rocks''.\\
	\texttt{geol\_permeability} & Surface permeability (log10).\\
	\bottomrule
\end{tabularx}
\end{table}

\section{Training Procedures}
\label{app:setup}

We train the EA-LSTM models based on the open-source code of \citet{Kratzert2019Benchmark} (Git version \texttt{2dd199e}, \url{https://github.com/kratzert/ealstm_regional_modeling}) for 30 epochs on NVIDIA P100 Pascal and V100 Volta GPUs using Python 3.7.3, PyTorch 1.1.0, and CUDA 9.0.
The initial learning rate of $0.001$ reduces to $0.0005$ after ten epochs and to $0.0001$ after another ten epochs.
We feed batches of 256 samples into the network, which consists of one 256-neuron hidden layer with a dropout rate of $0.4$.

For XGBoost (Git version \texttt{96cd7ec}, \url{https://github.com/dmlc/xgboost}) we use the same Python version and train on Intel Xeon E5-2683 v4 CPUs.
To find suitable hyperparameters, we perform two three-fold cross-validated random searches for each sequence length.
As we find that longer sequence lengths $k$ work better for larger datasets, we search on three years and 13 basins for sequence length 10, on six training years and 53 basins for sequence length 30, and on nine years and 265 basins for sequence length 100.
For each sequence length, both random searches fit up to \num{100} trees at a learning rate of $0.25$ (\texttt{learning\_rate}) and stop after \num{50} rounds without improvement (\texttt{early\_stopping\_rounds}).
First, we search for good tree parameters (\texttt{max\_depth}, \texttt{min\_child\_weight}, \texttt{colsample\_bytree}, \texttt{colsample\_bylevel}, \texttt{gamma}) in \num{5000} random samples.
Next, we use these parameters in a 100-iteration random search to identify regularization parameters (\texttt{reg\_alpha}, \texttt{reg\_lambda}).
Throughout the tuning procedure, we keep the row sampling (\texttt{subsample}) fixed at $0.9$.
\cref{tab:xgb-params} lists the final hyperparameters.
After parameter tuning, we train the XGBoost models with the final parameters on varying amounts of data at a learning rate of $0.08$ for up to \num{20000} iterations (\texttt{n\_estimators}); however, we stop once the NSE-loss on a validation set of $10\%$ of the training data does not improve for 100 rounds.

\begin{table}[pos=ht]
    \caption{Final XGBoost hyperparameters for each input sequence length $k$.
    We calibrate the hyperparameters for length 10 on three years and 13 basins, for length 30 on six years and 53 basins, and for length 100 on nine years and 265 basins.}
    \label{tab:xgb-params}
    \centering
    \begin{tabularx}{0.75\textwidth}{XS[table-format=5.3]S[table-format=5.3]S[table-format=5.3]}
        \toprule
        & \multicolumn{3}{c}{Input sequence length $k$} \\
        Parameter & {10} & {30} & {100} \\
        \midrule
        \texttt{n\_estimators} & 20000 & 20000 &  20000\\
        \texttt{early\_stopping\_rounds} & 100 & 100 & 100\\
        \texttt{learning\_rate} & 0.08 & 0.08 &  0.08\\
        \texttt{max\_depth} & 4 & 6 & 7\\
        \texttt{min\_child\_weight} & 1 & 1 & 9\\
        \texttt{colsample\_bytree} & 0.962 & 0.400 & 0.884\\
        \texttt{colsample\_bylevel} & 0.916 & 0.968 & 0.485\\
        \texttt{gamma} & 1.293 & 1.005 & 4.586\\
        \texttt{reg\_alpha} & 1.091 & 18.944 & 24.190\\
        \texttt{reg\_lambda} & 2.738 & 3.704 & 67.595\\
        \texttt{subsample} & 0.9 & 0.9 & 0.9\\
        \bottomrule
    \end{tabularx}
\end{table}

\section*{Acknowledgments}
This research was undertaken thanks in part to funding from the Canada First Research Excellence Fund, and enabled by computational resources provided by Compute Ontario and Compute Canada.
The authors further thank the Global Water Futures program and the Integrated Modeling Program for Canada (IMPC) for their financial support.

Our code, models, and results are publicly available at \url{https://github.com/gauchm/ealstm_regional_modeling}.
The CAMELS dataset is available at \url{https://ral.ucar.edu/solutions/products/camels}.

\bibliographystyle{cas-model2-names}
\bibliography{xgboost-camels}

\begin{thebibliography}{30}
\expandafter\ifx\csname natexlab\endcsname\relax\def\natexlab#1{#1}\fi
\providecommand{\url}[1]{\texttt{#1}}
\providecommand{\href}[2]{#2}
\providecommand{\path}[1]{#1}
\providecommand{\DOIprefix}{doi:}
\providecommand{\ArXivprefix}{arXiv:}
\providecommand{\URLprefix}{URL: }
\providecommand{\Pubmedprefix}{pmid:}
\providecommand{\doi}[1]{\href{http://dx.doi.org/#1}{\path{#1}}}
\providecommand{\Pubmed}[1]{\href{pmid:#1}{\path{#1}}}
\providecommand{\bibinfo}[2]{#2}
\ifx\xfnm\relax \def\xfnm[#1]{\unskip,\space#1}\fi
\bibitem[{Addor et~al.(2018)Addor, Nearing, Prieto, Newman, Le~Vine and
  Clark}]{Addor2018Ranking}
\bibinfo{author}{Addor, N.}, \bibinfo{author}{Nearing, G.},
  \bibinfo{author}{Prieto, C.}, \bibinfo{author}{Newman, A.J.},
  \bibinfo{author}{Le~Vine, N.}, \bibinfo{author}{Clark, M.P.},
  \bibinfo{year}{2018}.
\newblock \bibinfo{title}{A ranking of hydrological signatures based on their
  predictability in space}.
\newblock \bibinfo{journal}{Water Resources Research} \bibinfo{volume}{54},
  \bibinfo{pages}{8792--8812}.
\newblock \DOIprefix\doi{10.1029/2018WR022606}.
\bibitem[{Addor et~al.(2017)Addor, Newman, Mizukami and
  Clark}]{Addor2017CAMELS}
\bibinfo{author}{Addor, N.}, \bibinfo{author}{Newman, A.J.},
  \bibinfo{author}{Mizukami, N.}, \bibinfo{author}{Clark, M.P.},
  \bibinfo{year}{2017}.
\newblock \bibinfo{title}{The {CAMELS} data set: catchment attributes and
  meteorology for large-sample studies}.
\newblock \bibinfo{journal}{Hydrology and Earth System Sciences}
  \bibinfo{volume}{21}, \bibinfo{pages}{5293--5313}.
\newblock \DOIprefix\doi{10.5194/hess-21-5293-2017}.
\bibitem[{Amari(1993)}]{Amari1993Learning}
\bibinfo{author}{Amari, S.}, \bibinfo{year}{1993}.
\newblock \bibinfo{title}{A universal theorem on learning curves}.
\newblock \bibinfo{journal}{Neural Networks} \bibinfo{volume}{6},
  \bibinfo{pages}{161--166}.
\newblock \DOIprefix\doi{10.1016/0893-6080(93)90013-M}.
\bibitem[{Banko and Brill(2001)}]{Banko2001Size}
\bibinfo{author}{Banko, M.}, \bibinfo{author}{Brill, E.}, \bibinfo{year}{2001}.
\newblock \bibinfo{title}{Mitigating the paucity-of-data problem: Exploring the
  effect of training corpus size on classifier performance for natural language
  processing}, in: \bibinfo{booktitle}{Proceedings of the First International
  Conference on Human Language Technology Research},
  \bibinfo{publisher}{Association for Computational Linguistics},
  \bibinfo{address}{USA}. pp. \bibinfo{pages}{1--–5}.
\newblock \DOIprefix\doi{10.3115/1072133.1072204}.
\bibitem[{Best et~al.(2015)Best, Abramowitz, Johnson, Pitman, Balsamo, Boone,
  Cuntz, Decharme, Dirmeyer, Dong, Ek, Guo, Haverd, van~den Hurk, Nearing, Pak,
  Peters-Lidard, Santanello, Stevens and Vuichard}]{Best2015Benchmark}
\bibinfo{author}{Best, M.J.}, \bibinfo{author}{Abramowitz, G.},
  \bibinfo{author}{Johnson, H.R.}, \bibinfo{author}{Pitman, A.J.},
  \bibinfo{author}{Balsamo, G.}, \bibinfo{author}{Boone, A.},
  \bibinfo{author}{Cuntz, M.}, \bibinfo{author}{Decharme, B.},
  \bibinfo{author}{Dirmeyer, P.A.}, \bibinfo{author}{Dong, J.},
  \bibinfo{author}{Ek, M.}, \bibinfo{author}{Guo, Z.}, \bibinfo{author}{Haverd,
  V.}, \bibinfo{author}{van~den Hurk, B.J.J.}, \bibinfo{author}{Nearing, G.},
  \bibinfo{author}{Pak, B.}, \bibinfo{author}{Peters-Lidard, C.},
  \bibinfo{author}{Santanello, J.A.}, \bibinfo{author}{Stevens, L.},
  \bibinfo{author}{Vuichard, N.}, \bibinfo{year}{2015}.
\newblock \bibinfo{title}{The plumbing of land surface models: Benchmarking
  model performance}.
\newblock \bibinfo{journal}{Journal of Hydrometeorology} \bibinfo{volume}{16},
  \bibinfo{pages}{1425--1442}.
\newblock \DOIprefix\doi{10.1175/JHM-D-14-0158.1}.
\bibitem[{Chen and Guestrin(2016)}]{Chen2016XGB}
\bibinfo{author}{Chen, T.}, \bibinfo{author}{Guestrin, C.},
  \bibinfo{year}{2016}.
\newblock \bibinfo{title}{{XGBoost}: A scalable tree boosting system}, in:
  \bibinfo{booktitle}{Proceedings of the 22nd {ACM} {SIGKDD} International
  Conference on Knowledge Discovery and Data Mining}, \bibinfo{publisher}{ACM}.
  pp. \bibinfo{pages}{785--794}.
\newblock \DOIprefix\doi{10.1145/2939672.2939785}.
\bibitem[{Cohen(2013)}]{Cohen2013statistical}
\bibinfo{author}{Cohen, J.}, \bibinfo{year}{2013}.
\newblock \bibinfo{title}{Statistical power analysis for the behavioral
  sciences}.
\newblock \bibinfo{publisher}{Routledge}.
\bibitem[{Dastorani et~al.(2018)Dastorani, Mahjoobi, Talebi and
  Fakhar}]{Dastorani2018ML}
\bibinfo{author}{Dastorani, M.T.}, \bibinfo{author}{Mahjoobi, J.},
  \bibinfo{author}{Talebi, A.}, \bibinfo{author}{Fakhar, F.},
  \bibinfo{year}{2018}.
\newblock \bibinfo{title}{Application of machine learning approaches in
  rainfall-runoff modeling (case study:\ {Zayandeh Rood} basin in {Iran})}.
\newblock \bibinfo{journal}{Civil Engineering Infrastructures Journal}
  \bibinfo{volume}{51}, \bibinfo{pages}{293--310}.
\newblock \URLprefix \url{https://ceij.ut.ac.ir/article_68735.html},
  \DOIprefix\doi{10.7508/ceij.2018.02.004}.
\bibitem[{Dawson and Wilby(1998)}]{Dawson1998ANN}
\bibinfo{author}{Dawson, C.W.}, \bibinfo{author}{Wilby, R.},
  \bibinfo{year}{1998}.
\newblock \bibinfo{title}{An artificial neural network approach to
  rainfall-runoff modelling}.
\newblock \bibinfo{journal}{Hydrological Sciences Journal}
  \bibinfo{volume}{43}, \bibinfo{pages}{47--66}.
\newblock \DOIprefix\doi{10.1080/02626669809492102}.
\bibitem[{Dibike and Solomatine(2001)}]{Dibike2001ANN}
\bibinfo{author}{Dibike, Y.}, \bibinfo{author}{Solomatine, D.},
  \bibinfo{year}{2001}.
\newblock \bibinfo{title}{River flow forecasting using artificial neural
  networks}.
\newblock \bibinfo{journal}{Physics and Chemistry of the Earth, Part B:
  Hydrology, Oceans and Atmosphere} \bibinfo{volume}{26},
  \bibinfo{pages}{1--7}.
\newblock \DOIprefix\doi{10.1016/S1464-1909(01)85005-X}.
\bibitem[{Fekete et~al.(2015)Fekete, Robarts, Kumagai, Nachtnebel, Odada and
  Zhulidov}]{Fekete2015InSitu}
\bibinfo{author}{Fekete, B.M.}, \bibinfo{author}{Robarts, R.D.},
  \bibinfo{author}{Kumagai, M.}, \bibinfo{author}{Nachtnebel, H.P.},
  \bibinfo{author}{Odada, E.}, \bibinfo{author}{Zhulidov, A.V.},
  \bibinfo{year}{2015}.
\newblock \bibinfo{title}{Time for in situ renaissance}.
\newblock \bibinfo{journal}{Science} \bibinfo{volume}{349},
  \bibinfo{pages}{685--686}.
\newblock \DOIprefix\doi{10.1126/science.aac7358}.
\bibitem[{Friedman(2001)}]{Friedman2001Boosting}
\bibinfo{author}{Friedman, J.H.}, \bibinfo{year}{2001}.
\newblock \bibinfo{title}{Greedy function approximation: A gradient boosting
  machine}.
\newblock \bibinfo{journal}{The Annals of Statistics} \bibinfo{volume}{29},
  \bibinfo{pages}{1189--1232}.
\bibitem[{Gauch et~al.(2019a)Gauch, Gharari, Mai and Lin}]{Gauch2019Limited}
\bibinfo{author}{Gauch, M.}, \bibinfo{author}{Gharari, S.},
  \bibinfo{author}{Mai, J.}, \bibinfo{author}{Lin, J.}, \bibinfo{year}{2019}a.
\newblock \bibinfo{title}{Streamflow prediction with limited
  spatially-distributed input data}.
\newblock \bibinfo{journal}{Proceedings of the NeurIPS 2019 Workshop on
  Tackling Climate Change with Machine Learning} .
\bibitem[{Gauch et~al.(2019b)Gauch, Mai, Gharari and Lin}]{Gauch2019DataDriven}
\bibinfo{author}{Gauch, M.}, \bibinfo{author}{Mai, J.},
  \bibinfo{author}{Gharari, S.}, \bibinfo{author}{Lin, J.},
  \bibinfo{year}{2019}b.
\newblock \bibinfo{title}{Data-driven vs.\ physically-based streamflow
  prediction models}.
\newblock \bibinfo{journal}{Proceedings of 9th International Workshop on
  Climate Informatics} .
\bibitem[{Gauch et~al.(2019c)Gauch, Mai and Lin}]{Gauch2019Feeding}
\bibinfo{author}{Gauch, M.}, \bibinfo{author}{Mai, J.}, \bibinfo{author}{Lin,
  J.}, \bibinfo{year}{2019}c.
\newblock \bibinfo{title}{The proper care and feeding of {CAMELS}: How limited
  training data affects streamflow prediction}.
\newblock \href{http://arxiv.org/abs/1911.07249}{\tt arXiv:1911.07249}.
\bibitem[{Hestness et~al.(2017)Hestness, Narang, Ardalani, Diamos, Jun,
  Kianinejad, Patwary, Ali, Yang and Zhou}]{Hestness2017Scaling}
\bibinfo{author}{Hestness, J.}, \bibinfo{author}{Narang, S.},
  \bibinfo{author}{Ardalani, N.}, \bibinfo{author}{Diamos, G.},
  \bibinfo{author}{Jun, H.}, \bibinfo{author}{Kianinejad, H.},
  \bibinfo{author}{Patwary, M.}, \bibinfo{author}{Ali, M.},
  \bibinfo{author}{Yang, Y.}, \bibinfo{author}{Zhou, Y.}, \bibinfo{year}{2017}.
\newblock \bibinfo{title}{Deep learning scaling is predictable, empirically}.
\newblock \bibinfo{journal}{arXiv preprint arXiv:1712.00409} .
\bibitem[{Hochreiter and Schmidhuber(1997)}]{Hochreiter1997LSTM}
\bibinfo{author}{Hochreiter, S.}, \bibinfo{author}{Schmidhuber, J.},
  \bibinfo{year}{1997}.
\newblock \bibinfo{title}{Long short-term memory}.
\newblock \bibinfo{journal}{Neural computation} \bibinfo{volume}{9},
  \bibinfo{pages}{1735--1780}.
\bibitem[{Hu et~al.(2018)Hu, Wu, Li, Jian, Li and Lou}]{Hu2018DL}
\bibinfo{author}{Hu, C.}, \bibinfo{author}{Wu, Q.}, \bibinfo{author}{Li, H.},
  \bibinfo{author}{Jian, S.}, \bibinfo{author}{Li, N.}, \bibinfo{author}{Lou,
  Z.}, \bibinfo{year}{2018}.
\newblock \bibinfo{title}{Deep learning with a {Long Short-Term Memory}
  networks approach for rainfall-runoff simulation}.
\newblock \bibinfo{journal}{Water} \bibinfo{volume}{10}, \bibinfo{pages}{1543}.
\bibitem[{{Kaggle Team}(2015)}]{Kaggle2015Rossmann1}
\bibinfo{author}{{Kaggle Team}}, \bibinfo{year}{2015}.
\newblock \bibinfo{title}{Rossmann store sales, winner's interview: 1st place,
  {Gert Jacobusse}}.
\newblock \bibinfo{howpublished}{No Free Hunch}.
\newblock
  \bibinfo{note}{\url{http://blog.kaggle.com/2015/12/21/rossmann-store-sales-winners-interview-1st-place-gert/},
  accessed Nov 2019}.
\bibitem[{{Kaggle Team}(2016a)}]{Kaggle2016Grupo}
\bibinfo{author}{{Kaggle Team}}, \bibinfo{year}{2016}a.
\newblock \bibinfo{title}{{Grupo Bimbo} inventory demand, winners' interview:
  {Clustifier} \& {Alex} \& {Andrey}}.
\newblock \bibinfo{howpublished}{No Free Hunch}.
\newblock
  \bibinfo{note}{\url{http://blog.kaggle.com/2016/09/27/grupo-bimbo-inventory-demand-winners-interviewclustifier-alex-andrey/},
  accessed Nov 2019}.
\bibitem[{{Kaggle Team}(2016b)}]{Kaggle2016Rossmann2}
\bibinfo{author}{{Kaggle Team}}, \bibinfo{year}{2016}b.
\newblock \bibinfo{title}{Rossmann store sales, winner's interview: 2nd place,
  {Nima Shahbazi}}.
\newblock \bibinfo{howpublished}{No Free Hunch}.
\newblock
  \bibinfo{note}{\url{http://blog.kaggle.com/2016/02/03/rossmann-store-sales-winners-interview-2nd-place-nima-shahbazi/},
  accessed Nov 2019}.
\bibitem[{Kratzert et~al.(2018)Kratzert, Klotz, Brenner, Schulz and
  Herrnegger}]{Kratzert2018LSTM}
\bibinfo{author}{Kratzert, F.}, \bibinfo{author}{Klotz, D.},
  \bibinfo{author}{Brenner, C.}, \bibinfo{author}{Schulz, K.},
  \bibinfo{author}{Herrnegger, M.}, \bibinfo{year}{2018}.
\newblock \bibinfo{title}{Rainfall--runoff modelling using {Long Short-Term
  Memory} ({LSTM}) networks}.
\newblock \bibinfo{journal}{Hydrology and Earth System Sciences}
  \bibinfo{volume}{22}, \bibinfo{pages}{6005--6022}.
\newblock \DOIprefix\doi{10.5194/hess-22-6005-2018}.
\bibitem[{Kratzert et~al.(2019)Kratzert, Klotz, Shalev, Klambauer, Hochreiter
  and Nearing}]{Kratzert2019Benchmark}
\bibinfo{author}{Kratzert, F.}, \bibinfo{author}{Klotz, D.},
  \bibinfo{author}{Shalev, G.}, \bibinfo{author}{Klambauer, G.},
  \bibinfo{author}{Hochreiter, S.}, \bibinfo{author}{Nearing, G.},
  \bibinfo{year}{2019}.
\newblock \bibinfo{title}{Benchmarking a catchment-aware {Long Short-Term
  Memory} network ({LSTM}) for large-scale hydrological modeling}.
\newblock \bibinfo{journal}{Hydrology and Earth System Sciences Discussions} ,
  \bibinfo{pages}{1--32}\DOIprefix\doi{10.5194/hess-2019-368}.
\bibitem[{Luo et~al.(2019)Luo, Huang, Hu, Li and Zhang}]{Luo2019KDD}
\bibinfo{author}{Luo, Z.}, \bibinfo{author}{Huang, J.}, \bibinfo{author}{Hu,
  K.}, \bibinfo{author}{Li, X.}, \bibinfo{author}{Zhang, P.},
  \bibinfo{year}{2019}.
\newblock \bibinfo{title}{{AccuAir}: Winning solution to air quality prediction
  for {KDD Cup} 2018}, in: \bibinfo{booktitle}{Proceedings of the 25th {ACM}
  {SIGKDD} International Conference on Knowledge Discovery \& Data Mining},
  \bibinfo{publisher}{ACM}. pp. \bibinfo{pages}{1842--1850}.
\newblock \DOIprefix\doi{10.1145/3292500.3330787}.
\bibitem[{Maurer et~al.(2002)Maurer, Wood, Adam, Lettenmaier and
  Nijssen}]{Maurer2002Forcings}
\bibinfo{author}{Maurer, E.P.}, \bibinfo{author}{Wood, A.W.},
  \bibinfo{author}{Adam, J.C.}, \bibinfo{author}{Lettenmaier, D.P.},
  \bibinfo{author}{Nijssen, B.}, \bibinfo{year}{2002}.
\newblock \bibinfo{title}{A long-term hydrologically based dataset of land
  surface fluxes and states for the conterminous {United States}}.
\newblock \bibinfo{journal}{Journal of Climate} \bibinfo{volume}{15},
  \bibinfo{pages}{3237--3251}.
\newblock \DOIprefix\doi{10.1175/1520-0442(2002)015<3237:ALTHBD>2.0.CO;2}.
\bibitem[{Mittelhammer et~al.(2000)Mittelhammer, Judge and
  Miller}]{Mittelhammer2000Econometric}
\bibinfo{author}{Mittelhammer, R.C.}, \bibinfo{author}{Judge, G.G.},
  \bibinfo{author}{Miller, D.J.}, \bibinfo{year}{2000}.
\newblock \bibinfo{title}{Econometric Foundations}.
  \bibinfo{publisher}{Cambridge University Press}.
\newblock pp. \bibinfo{pages}{73--74}.
\bibitem[{Mizukami et~al.(2019)Mizukami, Rakovec, Newman, Clark, Wood, Gupta
  and Kumar}]{Mizukami2019Metrics}
\bibinfo{author}{Mizukami, N.}, \bibinfo{author}{Rakovec, O.},
  \bibinfo{author}{Newman, A.J.}, \bibinfo{author}{Clark, M.P.},
  \bibinfo{author}{Wood, A.W.}, \bibinfo{author}{Gupta, H.V.},
  \bibinfo{author}{Kumar, R.}, \bibinfo{year}{2019}.
\newblock \bibinfo{title}{On the choice of calibration metrics for
  ``high-flow'' estimation using hydrologic models}.
\newblock \bibinfo{journal}{Hydrology and Earth System Sciences}
  \bibinfo{volume}{23}, \bibinfo{pages}{2601--2614}.
\newblock \DOIprefix\doi{10.5194/hess-23-2601-2019}.
\bibitem[{Newman et~al.(2017)Newman, Mizukami, Clark, Wood, Nijssen and
  Nearing}]{Newman2017Benchmark}
\bibinfo{author}{Newman, A.J.}, \bibinfo{author}{Mizukami, N.},
  \bibinfo{author}{Clark, M.P.}, \bibinfo{author}{Wood, A.W.},
  \bibinfo{author}{Nijssen, B.}, \bibinfo{author}{Nearing, G.},
  \bibinfo{year}{2017}.
\newblock \bibinfo{title}{Benchmarking of a physically based hydrologic model}.
\newblock \bibinfo{journal}{Journal of Hydrometeorology} \bibinfo{volume}{18},
  \bibinfo{pages}{2215--2225}.
\newblock \DOIprefix\doi{10.1175/JHM-D-16-0284.1},
  \href{http://arxiv.org/abs/https://doi.org/10.1175/JHM-D-16-0284.1}{\tt
  arXiv:https://doi.org/10.1175/JHM-D-16-0284.1}.
\bibitem[{Newman et~al.(2014)Newman, Sampson, Clark, Bock, Viger and
  Blodgett}]{Newman2014Camels}
\bibinfo{author}{Newman, A.J.}, \bibinfo{author}{Sampson, K.},
  \bibinfo{author}{Clark, M.P.}, \bibinfo{author}{Bock, A.},
  \bibinfo{author}{Viger, R.}, \bibinfo{author}{Blodgett, D.},
  \bibinfo{year}{2014}.
\newblock \bibinfo{title}{A large-sample watershed-scale hydrometeorological
  dataset for the contiguous {USA}}.
\newblock \DOIprefix\doi{10.5065/d6mw2f4d}.
\bibitem[{Sawilowsky(2009)}]{Sawilowsky2009Effect}
\bibinfo{author}{Sawilowsky, S.S.}, \bibinfo{year}{2009}.
\newblock \bibinfo{title}{New effect size rules of thumb}.
\newblock \bibinfo{journal}{Journal of Modern Applied Statistical Methods}
  \bibinfo{volume}{8}, \bibinfo{pages}{597--599}.
\newblock \DOIprefix\doi{10.22237/jmasm/1257035100}.

\end{thebibliography}

\end{document}